\documentclass[11pt]{article}

\usepackage{setspace}
\usepackage{amssymb}
\usepackage{amsmath}
\usepackage[T1]{fontenc}
\usepackage{makecell}
\usepackage{graphics}
\usepackage{float}
\usepackage{setspace}
\usepackage{pgfplots}
\usepackage{authblk}

\title{Multi-view Low-rank Sparse Subspace Clustering}

\author{Maria Brbi\'c \footnote{maria.brbic@irb.hr}}
\author{Ivica Kopriva \footnote{ivica.kopriva@irb.hr}}

\affil{Rudjer Boskovic Institute, Zagreb, Croatia}

\date{August 2017}

\usepackage{hyperref}
\hypersetup{
    bookmarks=true,         % show bookmarks bar?
    unicode=false,          % non-Latin characters in Acrobat‚Äôs bookmarks
    colorlinks=true,       % false: boxed links; true: colored links
    linkcolor=blue,          % color of internal links
    urlcolor=blue           % color of external links
    citecolor=green,        % color of links to bibliography
}

\begin{document}

\maketitle

\begin{abstract}

Most existing approaches address multi-view subspace clustering problem by constructing the affinity matrix on each view separately and afterwards propose how to extend spectral clustering algorithm to handle multi-view data. This paper presents an approach to multi-view subspace clustering that learns a joint subspace representation by constructing affinity matrix shared among all views. Relying on the importance of both low-rank and sparsity constraints in the construction of the affinity matrix, we introduce the objective that balances between the agreement across different views, while at the same time encourages sparsity and low-rankness of the solution. Related low-rank and sparsity constrained optimization problem is for each view solved using the alternating direction method of multipliers. Furthermore, we extend our approach to cluster data drawn from nonlinear subspaces by solving the corresponding problem in a reproducing kernel Hilbert space. The proposed algorithm outperforms state-of-the-art multi-view subspace clustering algorithms on one synthetic and four real-world datasets.

\end{abstract}

\section{Introduction}
In many real-world machine learning problems the same data is comprised of several different representations or views. For example, same documents may be available in multiple languages \cite{Amini09} or different descriptors can be constructed from the same images \cite{Vedaldi09}. Although each of these individual views may be sufficient to perform a learning task, integrating complementary information from different views can reduce the complexity of a given task \cite{Xu13survey}. Multi-view clustering seeks to partition data points based on multiple representations by assuming that the same cluster structure is shared across views. By combining information from different views, multi-view clustering algorithms attempt to achieve more accurate cluster assignments than one can get by simply concatenating features from different views.

In practice, high-dimensional data often reside in a low-dimensional subspace. When all data points lie in a single subspace, the problem can be set as finding a basis of a subspace and a low-dimensional representation of data points. Depending on the constraints imposed on the low-dimensional representation, this problem can be solved using e.g. Principal Component Analysis (PCA) \cite{Jolliffe1986}, Independent Component Analysis (ICA) \cite{Comon1994} or Non-negative Matrix Factorization (NMF) \cite{Lee99, LiuTao16, Liu13NMF}. On the other hand, data points can be drawn from different sources and lie in a union of subspaces. By assigning each subspace to one cluster, one can solve the problem by applying standard clustering algorithms, such as k-means \cite{Macqueen67}. However, these algorithms are based on the assumption that data points are distributed around centroid and often do not perform well in the cases when data points in a subspace are arbitrarily distributed. %two points can be very close and lie in different subspace and can be far and lie in same subspace
For example, two points can have a small distance and lie in different subspaces or can be far and still lie in the same subspace \cite{Vidal11}. Therefore, methods that rely on a spatial proximity of data points often fail to provide a satisfactory solution.
This has motivated the development of subspace clustering algorithms \cite{Vidal11}. The goal of subspace clustering is to identify the low-dimensional subspaces and find the cluster membership of data points. Spectral based methods \cite{Shi2000, Ng01,Filippone08} present one approach to subspace clustering problem. They have gained a lot of attention in the recent years due to the competitive results they achieve on arbitrarily shaped clusters and their well defined mathematical principles. These methods are based on the spectral graph theory and represent data points as nodes in a weighted graph. The clustering problem is then solved as a relaxation of the min-cut problem on a graph \cite{vonLuxburg2007}. 

One of the main challenges in spectral based methods is the construction of the affinity matrix whose elements define the similarity between data points. Sparse subspace clustering \cite{Elhamifar13} and low-rank subspace clustering \cite{Liu10, Liu13,Favaro11, Vidal14} are among most effective methods that solve this problem. These methods rely on the self-expressiveness property of the data by representing each data point as a linear combination of other data points. Low-Rank Representation (LRR) \cite{Liu10, Liu13} imposes low-rank constraint on the data representation matrix and captures global structure of the data. Low-rank implies that data matrix is represented by a sum of small number of outer products of left and right singular vectors weighted by corresponding singular values. Under assumption that subspaces are independent and data sampling is sufficient, LRR guarantees exact clustering. However, for many real-world datasets this assumption is overly restrictive and the assumption that data is drawn from disjoint subspaces would be more appropriate \cite{Tang14, Tang16}. On the other hand, Sparse Subspace Clustering (SSC) \cite{Elhamifar13} represents each data point as a sparse linear combination of other points and captures local structure of the data. Learning representation matrix in SSC can be interpreted as sparse coding \cite{Olshausen96, Aharon06, Elad06,Mairal08, Mairal10, Filipovic11}. However, compared to sparse coding where dictionary is learned such that the representation is sparse \cite{LiuZha16, YuRui11}, SSC is based on self-representation property i.e. data matrix stands for a dictionary. SSC also succeeds when data is drawn from independent subspaces and the conditions have been established for clustering data drawn from disjoint subspaces \cite{Elhamifar10}. However, theoretical analysis in \cite{Nasihatkon11} shows that it is possible that SSC over-segments subspaces when the dimensionality of data points is higher than three. Experimental results in \cite{Wang13} show that LRR misclassifies different data points than SSC. Therefore, in order to capture global and the local structure of the data, it is necessary to combine low-rank and sparsity constraints \cite{Wang13, Liu2016}.

Multi-view subspace clustering can be considered as a part of multi-view or multi-modal learning. Multi-view learning method in \cite{Xu15} learns view generation matrices and representation matrix, relying on the assumption that data from all the views share the same representation matrix. The multi-view method in \cite{Yang17} is based on the canonical correlation analysis in extraction of two-view filter-bank-based features for image classification task. Similarly, in \cite{Luo15a} the authors rely on tensor-based canonical correlation analysis to perform multi-view dimensionality reduction. This approach can be used as a preprocessing step in multi-view learning in case of high-dimensional data. In \cite{Hong15} low-rank representation matrix is learned on each view separately and learned representation matrices are concatenated to a matrix from which a unified graph affinity matrix is obtained. The method in \cite{Yu17} relies on learning a linear projection matrix for each view separately. High-order distance-based multi-view stochastic learning is proposed in \cite{Yu14a}, to efficiently explore the complementary characteristics of multi-view features for image classification. The method in \cite{Yu14b} is application oriented towards image reranking and assumes that multi-view features are contained in hypergraph Laplacians that define different modalities. In \cite{Luo15b} authors propose multi-view matrix completion algorithm for handling multi-view features in semi-supervised multi-label image classification. 

Previous multi-view subspace clustering works \cite{Kumar11, Cheng11, Xia14, Lu16} address the problem by constructing affinity matrix on each view separately and then extend algorithm to handle multi-view data. However, since input data may often be corrupted by noise, this approach can lead to the propagation of noise in the affinity matrices and degrade clustering performance. Different from the existing approaches, we propose multi-view spectral clustering framework that jointly learns a subspace representation by constructing single affinity matrix shared by multi-view data, while at the same time encourages low-rank and sparsity of the representation. We propose Multi-view Low-rank Sparse Subspace Clustering (MLRSSC) algorithms that enforce agreement: (i) between affinity matrices of the pairs of views; (ii) between affinity matrices towards a common centroid. Opposed to \cite{Yang17, Yu14b, Yu17b}, the proposed approach can deal with highly heterogeneous multi-view data coming from different modalities. We present optimization procedure to solve the convex dual optimization problems using Alternating Direction Method of Multipliers (ADMM) \cite{Boyd11}.
Furthermore, we propose the kernel extension of our algorithms by solving the problem in a Reproducing Kernel Hilbert Space (RKHS). Experimental results show that MLRSSC algorithm outperforms state-of-the-art multi-view subspace clustering algorithms on several benchmark datasets. Additionally, we evaluate performance on a novel real-world heterogeneous multi-view dataset from biological domain.

The remainder of the paper is organized as follows. Section 2 gives a brief overview of the low-rank and sparse subspace clustering methods. Section 3 introduces two novel multi-view subspace clustering algorithms. In Section 4 we present the kernelized version of the proposed algorithms by formulating subspace clustering problem in RKHS. The performance of the new algorithms is demonstrated in Section 5. Section 6 concludes the paper. 

\section{Background and Related Work}

In this section, we give a brief introduction to Sparse Subspace Clustering (SSC) \cite{Elhamifar13}, Low-Rank Representation (LRR) \cite{Liu10, Liu13} and Low-rank Sparse Subspace Clustering (LRSSC) \cite{Wang13}.

 \subsection{Main Notations}
Throughout this paper, matrices are represented with bold capital symbols and vectors with bold lower-case symbols. $\|\cdot\|_F$ denotes the Frobenius norm of a matrix. The $\ell_1$ norm, denoted by $\|\cdot\|_1$, is the sum of absolute values of matrix elements; infinity norm $\|\cdot\|_\infty$ is the maximum absolute element value; and the nuclear norm $\|\cdot\|_*$ is the sum of singular values of a matrix. Trace operator of a matrix is denoted by $tr(\cdot)$ and $diag(\cdot)$ is the vector of diagonal elements of a matrix. $\mathbf{0}$ denotes null vector. Table \ref{notation_summary} summarizes some notations used throughout the paper.

\begin{table}[t]
\caption{Notations and abbreviations}
\fontsize{9}{11}\selectfont
\begin{center} \label{notation_summary} 
\begin{tabular}{ l  l  } 
\hline 
 \textbf{Notation} & \textbf{Definition} \\ 
 \hline
$N$ & Number of data points  \\ 
 $k$ & Number of clusters  \\ 
 $v$ & View index \\ 
 $n_v$ & Number of views \\ 
 $D^{(v)}$ & Dimension of data points in a view $v$ \\ 
 $\mathbf{X}^{(v)}\in {\rm I\!R}^{D^{(v)}\times N}$ & Data matrix in a view $v$   \\ 
 $\mathbf{C}^{(v)}\in {\rm I\!R}^{N\times N}$ & Representation matrix in a view $v$   \\ 
  $\mathbf{C}^{*}\in {\rm I\!R}^{N\times N}$ & Centroid representation matrix   \\ 
 $\mathbf{W}\in {\rm I\!R}^{N\times N}$ & Affinity matrix   \\ 
  $\mathbf{X}=\mathbf{U\Sigma} \mathbf{V}^T$ & Singular value decomposition (SVD) of $\mathbf{X}$   \\
  $\Phi(\mathbf{X}^{(v)})$ & Data points in a view $v$ mapped into high-dimensional feature space\\
  $\mathbf{K}^{(v)}\in {\rm I\!R}^{N\times N}$ & Gram matrix in a view $v$\\
 \hline
\end{tabular}
%\\[8pt]
\end{center}
\end{table}

\subsection{Related Work}
Consider the set of $N$ data points $\mathbf{X}=\big\{\mathbf{x}_i\in  {\rm I\!R}^D\big\}_{i=1}^N$ that lie in a union of $k>1$ linear subspaces of unknown dimensions. Given the set of data points $\mathbf{X}$, the task of subspace clustering is to cluster data points according to the subspaces they belong to. The first step is the construction of the affinity matrix $\mathbf{W} \in {\rm I\!R}^{N\times N}$ whose elements define the similarity between data points. Ideally, the affinity matrix is a block diagonal matrix such that a nonzero distance is assigned to the points from the same subspace. LRR, SSC and LRSSC construct the affinity matrix by enforcing low-rank, sparsity and low-rank plus sparsity constraints, respectively.

Low-Rank Representation (LRR) \cite{Liu10, Liu13} seeks to find a low-rank representation matrix $\mathbf{C}\in {\rm I\!R}^{N\times N}$ for input data $\mathbf{X}$. The basic model of LRR is the following:
\begin{equation} \label{LRR}
\min_\mathbf{C}\big\|\mathbf{C}\big\|_*\ \  s.t.\ \ \mathbf{X}=\mathbf{XC},
\end{equation}
where the nuclear norm is used to approximate the rank of $\mathbf{C}$ and that results in the convex optimization problem.

Denote the SVD of $\mathbf{X}$ as $\mathbf{U\Sigma V}^T$. The minimizer of equation (\ref{LRR}) is uniquely given by \cite{Liu10}:
\begin{equation}
\hat{\mathbf{C}}=\mathbf{V}\mathbf{V}^T.
\end{equation}
In the cases when data is contaminated by noise, the following problem needs to be solved:
\begin{equation} \label{LRR_noise}
\min_\mathbf{C}\frac{1}{2}\big\|\mathbf{X}-\mathbf{XC}\big\|_F^2+\lambda\big\|\mathbf{C}\big\|_*.
\end{equation}
The optimal solution of equation (\ref{LRR_noise}) has been derived in \cite{Favaro11}:
\begin{equation}
\hat{\mathbf{C}}=\mathbf{V}_1(\mathbf{I}-\frac{1}{\lambda}\mathbf{\Sigma}_1^{-2})\mathbf{V_1}^T,
\end{equation}
where $\mathbf{U}=[\mathbf{U}_1\ \mathbf{U}_2]$, $\mathbf{\Sigma}=diag(\mathbf{\Sigma}_1\ \mathbf{\Sigma}_2)$ and $\mathbf{V}=[\mathbf{V}_1\ \mathbf{V}_2]$. Matrices are partitioned according to the sets $\mathcal{I}_1=\{i:\sigma_i>\frac{1}{\sqrt\lambda}\}$ and $\mathcal{I}_2=\{i:\sigma_i\leq\frac{1}{\sqrt\lambda}\}$.

Sparse Subspace Clustering (SSC) \cite{Elhamifar13} requires that each data point is represented by a small number of data points from its own subspace and it amounts to solve the following minimization problem:
\begin{equation}
\min_\mathbf{C}\big\|\mathbf{C}\big\|_1\ \  s.t.\ \ \mathbf{X}=\mathbf{XC},\ diag(\mathbf{C})=\mathbf{0}.
\end{equation} 
The $\ell_1$ norm is used as the tightest convex relaxation of the $\ell_0$ quasi-norm that counts the number of nonzero elements of the solution. Constraint $diag(\mathbf{C})=\mathbf{0}$ is used to avoid trivial solution of representing a data point as a linear combination of itself.

If data is contaminated by noise, the following minimization problem needs to be solved:
\begin{equation}
\min_\mathbf{C}\frac{1}{2}\big\|\mathbf{X}-\mathbf{XC}\big\|_F^2+\lambda\big\|\mathbf{C}\big\|_1\ \  s.t.\ \ diag(\mathbf{C})=\mathbf{0}.
\end{equation}
This problem can be efficiently solved using ADMM optimization procedure \cite{Boyd11}.

Low-Rank Sparse Subspace Clustering (LRSSC) \cite{Wang13} combines low-rank and sparsity constraints:
\begin{equation}
\min_\mathbf{C}\big\|\mathbf{C}\big\|_*+\lambda\big\|\mathbf{C}\big\|_1\ \  s.t.\ \  \mathbf{X}=\mathbf{XC},\ diag(\mathbf{C})=\mathbf{0}.
\end{equation}
In the case of the corrupted data the following problem needs to be solved to approximate $\mathbf{C}$:
\begin{equation}
\min_\mathbf{C}\frac{1}{2}\big\|\mathbf{X}-\mathbf{XC}\big\|_F^2+\beta_1\big\|\mathbf{C}\big\|_*+\beta_2\big\|\mathbf{C}\big\|_1\ \ s.t.\ \  diag(\mathbf{C})=\mathbf{0}.
\end{equation}
Once matrix $\mathbf{C}$ is obtained by LRR, SSC or LRSSC approach, the affinity matrix $\mathbf{W}$ is calculated as:
\begin{equation}
\mathbf{W}=|\mathbf{C}|+|\mathbf{C}|^T.
\end{equation}
Given affinity matrix $\mathbf{W}$, spectral clustering \cite{Shi2000, Ng01} finds cluster membership of data points by applying k-means clustering to the eigenvectors of the graph Laplacian matrix $\mathbf{L}\in {\rm I\!R}^{N\times N}$ computed from the affinity matrix $\mathbf{W}$. 

\section{Multi-view Low-rank Sparse Subspace Clustering}

In this section we present Multi-view Low-rank Sparse Subspace Clustering (MLRSSC) algorithm with two different regularization approaches. 
We assume that we are given a dataset $\mathbf{X}=\big\{\mathbf{X}^{(1)},\mathbf{X}^{(2)},\\...,\mathbf{X}^{(n_v)}\big\}$ of $n_v$ views, where each $\mathbf{X}^{(i)}=\big\{\mathbf{x}_j^{(i)}\in {\rm I\!R}^{D^{(i)}}\big\}_{j=1}^N$ is described with its own set of $D^{(i)}$ features. Our objective is to find a joint representation matrix $\mathbf{C}$ that balances trade-off between the agreement across different views, while at the same time promotes sparsity and low-rankness of the solution. 

 We formulate joint objective function that enforces representation matrices $\allowbreak\big\{\mathbf{C}^{(1)},\mathbf{C}^{(2)},...,\mathbf{C}^{(n_v)}\big\}$ across different views to be regularized towards a common consensus. Motivated by \cite{Kumar11}, we propose two regularization schemes of the MLRSSC algorithm: (i) MLRSSC based on pairwise similarities and (ii) centroid-based MLRSSC. The first regularization encourages similarity between pairs of representation matrices. The centroid-based approach enforces representations across different views towards a common centroid. Standard spectral clustering algorithm can then be applied to the jointly inferred affinity matrix.

\subsection{Pairwise Multi-view Low-rank Sparse Subspace Clustering}

We propose to solve the following joint optimization problem over $n_v$ views:
\begin{equation} \label{obj_funct_all_pair}
\begin{split}
\min_{\mathbf{C}^{(1)},\mathbf{C}^{(2)},...,\mathbf{C}^{(n_v)}}& \sum_{v=1}^{n_v} \big(\beta_1\big\|\mathbf{C}^{(v)}\big\|_*+\beta_2\big\|\mathbf{C}^{(v)}\big\|_1\big)+\sum_{1\leq v,w\leq n_v, v\neq w}\lambda^{(v)}\big\|\mathbf{C}^{(v)}-\mathbf{C}^{(w)}\big\|_F^2
\\ & s.t.\quad \mathbf{X}^{(v)}=\mathbf{X}^{(v)}\mathbf{C}^{(v)},\quad diag(\mathbf{C}^{(v)})=\mathbf{0},\quad v=1,...n_v,
\end{split}
\end{equation}
where $\mathbf{C}^{(v)}\in  {\rm I\!R}^{N\times N} $ is the representation matrix for view $v$. Parameters $\beta_1, \beta_2$ and $\lambda^{(v)}$ define the trade-off between low-rank, sparsity constraint and the agreement across views, respectively. In the cases where we do not have a prior information that one view is more important than the others, $\lambda^{(v)}$ does not dependent on a view $v$ and the same value of $\lambda^{(v)}$ is used across all views $v=1,...,n_v$. The last term in the objective in (\ref{obj_funct_all_pair}) is introduced to encourage similarities between pairs of representation matrices across views.

With all but one $\mathbf{C}^{(v)}$ fixed, we minimize the function (\ref{obj_funct_all_pair}) for each $\mathbf{C}^{(v)}$ independently:
\begin{equation} \label{obj_funct_pair}
\begin{split}
\min_{\mathbf{C}^{(v)}} \beta_1\big\|\mathbf{C}^{(v)}\big\|_*&+\beta_2\big\|\mathbf{C}^{(v)}\big\|_1+\lambda^{(v)}\sum_{1\leq w\leq n_v, v\neq w}\big\|\mathbf{C}^{(v)}-\mathbf{C}^{(w)}\big\|_F^2 
\\ & s.t.\quad \mathbf{X}^{(v)}=\mathbf{X}^{(v)}\mathbf{C}^{(v)},\quad diag(\mathbf{C}^{(v)})=\mathbf{0}.
\end{split}
\end{equation}
By introducing auxiliary variables $\mathbf{C}_1^{(v)}$, $\mathbf{C}_2^{(v)}$, $\mathbf{C}_3^{(v)}$ and $\mathbf{A}^{(v)}$, we reformulate the objective:
\begin{equation} \label{min_aux}
\begin{split}
&\min_{\mathbf{C}_1^{(v)},\mathbf{C}_2^{(v)},\mathbf{C}_3^{(v)},\mathbf{A}^{(v)}} \beta_1\big\|\mathbf{C}^{(v)}_1\big\|_*+\beta_2\big\|\mathbf{C}^{(v)}_2\big\|_1+\lambda^{(v)}\sum_{1\leq w\leq n_v, v\neq w}\big\|\mathbf{C}^{(v)}_3-\mathbf{C}^{(w)}\big\|_F^2 
\\  s.t.&\quad \mathbf{X}^{(v)}=\mathbf{X}^{(v)}\mathbf{A}^{(v)},\quad \mathbf{A}^{(v)}=\mathbf{C}_2^{(v)}-diag(\mathbf{C}_2^{(v)}),\quad \mathbf{A}^{(v)}=\mathbf{C}_1^{(v)}, \quad \mathbf{A}^{(v)}=\mathbf{C}_3^{(v)}.
\end{split}
\end{equation}
The augmented Lagrangian is:
\begin{equation} \label{Lag}
\begin{split}
\mathcal{L}&\big({\{{\mathbf{C}_i^{(v)}}\}}_{i=1}^3,\mathbf{A}^{(v)},{\{\mathbf{\Lambda}_i^{(v)}\}}_{i=1}^4\big)=  \beta_1\big\|\mathbf{C}_1^{(v)}\big\|_*+\beta_2\big\|\mathbf{C}_2^{(v)}\big\|_1\\&+\lambda^{(v)}\sum_{1\leq w\leq n_v, w\neq v}\big\|\mathbf{C}_3^{(v)}-\mathbf{C}^{(w)}\big\|_F^2+ \frac{\mu_1}{2}\big\|\mathbf{X}^{(v)}-\mathbf{X}^{(v)}\mathbf{A}^{(v)}\big\|_F^2
\\&+\frac{\mu_2}{2}\big\|\mathbf{A}^{(v)}-\mathbf{C}_2^{(v)}+diag(\mathbf{C}_2^{(v)})\big\|_F^2+\frac{\mu_3}{2}\big\|\mathbf{A}^{(v)}-\mathbf{C}_1^{(v)}\big\|_F^2
+\frac{\mu_4}{2}\big\|\mathbf{A}^{(v)}-\mathbf{C}_3^{(v)}\big\|_F^2
\\&+tr\big[{\mathbf{\Lambda}_1^{(v)}}^T\big(\mathbf{X}^{(v)}-\mathbf{X}^{(v)}\mathbf{A}^{(v)}\big)\big]
+tr\big[{\mathbf{\Lambda}_2^{(v)}}^T\big(\mathbf{A}^{(v)}-\mathbf{C}_2^{(v)}+diag(\mathbf{C}_2^{(v)})\big)\Big]
\\&+tr\big[{\mathbf{\Lambda}_3^{(v)}}^T\big(\mathbf{A}^{(v)}-\mathbf{C}_1^{(v)}\big)\big]+tr\big[{\mathbf{\Lambda}_4^{(v)}}^T\big(\mathbf{A}^{(v)}-\mathbf{C}_3^{(v)}\big)\big],
\end{split}
\end{equation}
where ${\{\mu_i>0\}}_{i=1}^3$ are penalty parameters that need to be tuned and ${\{\mathbf{\Lambda}_i^{(v)}\}}_{i=1}^4$ are Lagrange dual variables.

To solve the convex optimization problem in (\ref{min_aux}), we use Alternating Direction Method of Multipliers (ADMM) \cite{Boyd11}. ADMM converges for the objective composed of two-block convex separable problems, but here the terms $\mathbf{C}_1^{(v)}$, $\mathbf{C}_2^{(v)}$ and $\mathbf{C}_3^{(v)}$ do not depend on each other and can be observed as one variable block.

\textbf{Update rule for $\mathbf{A}^{(v)}$ at iteration $k+1$}. Given ${\{\mathbf{C}_i^{(v)}\}}_{i=1}^3, {\{\mathbf{\Lambda}_i^{(v)}\}}_{i=1}^4$ at iteration $k$, the matrix $\mathbf{A}^{(v)}$ that minimizes the objective in equation (\ref{Lag}) is updated by the following update rule:
\begin{equation} \label{A_pairwise_update}
\begin{split}
\mathbf{A}^{(v)}=&\big[\mu_1{\mathbf{X}^{(v)}}^T{\mathbf{X}^{(v)}}+(\mu_2+\mu_3+\mu_4)\mathbf{I}\big]^{-1}\times\big(\mu_1{\mathbf{X}^{(v)}}^T{\mathbf{X}^{(v)}}+\mu_2\mathbf{C}_2^{(v)}\\&+\mu_3\mathbf{C}_1^{(v)}+\mu_4\mathbf{C}_3^{(v)}+{\mathbf{X}^{(v)}}^T{\mathbf{\Lambda}_1^{(v)}}-{\mathbf{\Lambda}_2^{(v)}}-{\mathbf{\Lambda}_3^{(v)}}-{\mathbf{\Lambda}_4^{(v)}}\big).
\end{split}
\end{equation}
The update rule follows straightforwardly by setting the partial derivative of $\mathcal{L}$ in equation (\ref{Lag}) with respect to $\mathbf{A}^{(v)}$ to zero.

\textbf{Update rule for $\mathbf{C}_1^{(v)}$ at iteration $k+1$.} Given $ \mathbf{A}^{(v)}$ at iteration $k+1$ and $\mathbf{\Lambda}_3^{(v)}$ at iteration $k$, we minimize the objective in equation (\ref{Lag}) with respect to $\mathbf{C}_1^{(v)}$:
\begin{equation} \label{Lag_C1}
\begin{split}
\min_{\mathbf{C}_1^{(v)}}&\mathcal{L}\big(\mathbf{C}_1^{(v)},\mathbf{A}^{(v)},{\mathbf{\Lambda}_3^{(v)}}\big)\\&=\min_{\mathbf{C}_1^{(v)}}\beta_1\big\|\mathbf{C}_1^{(v)}\big\|_*+\frac{\mu_3}{2}\big\|\mathbf{A}^{(v)}-\mathbf{C}_1^{(v)}\big\|_F^2
+tr\big[{\mathbf{\Lambda}_3^{(v)}}^T\big(\mathbf{A}^{(v)}-\mathbf{C}_1^{(v)}\big)\big] \\
&=\min_{\mathbf{C}_1^{(v)}}\beta_1\big\|\mathbf{C}_1^{(v)}\big\|_*+\frac{\mu_3}{2}\big\|\mathbf{A}^{(v)}-\mathbf{C}_1^{(v)}\big\|_F^2+tr\big[{\mathbf{\Lambda}_3^{(v)}}^T\big(\mathbf{A}^{(v)}-\mathbf{C}_1^{(v)}\big)\big]+\frac{\big\|{\mathbf{\Lambda}_3^{(v)}}\big\|_F^2}{2\mu_3}\\
 &=\min_{\mathbf{C}_1^{(v)}}\beta_1\big\|\mathbf{C}_1^{(v)}\big\|_*+\frac{\mu_3}{2}\Big\|\mathbf{A}^{(v)}-\mathbf{C}_1^{(v)}+\frac{\mathbf{\Lambda}_3^{(v)}}{\mu_3}\Big\|_F^2,
\end{split}
\end{equation}
From \cite{Cai10}, it follows that the the unique minimizer of (\ref{Lag_C1}) is:
\begin{equation} \label{C1_pairwise_update}
\mathbf{C}_1^{(v)}=\Pi_\frac{\beta_1}{\mu_3}\Big(\mathbf{A}^{(v)}+\frac{{\mathbf{\Lambda}_3^{(v)}}}{\mu_3}\Big),
\end{equation}
where $\Pi_\beta(\mathbf{Y})=\mathbf{U}\pi_\beta(\mathbf{\Sigma})\mathbf{V}^T$ performs soft-thresholding operation on the singular values of $\mathbf{Y}$ and $\mathbf{U}\mathbf{\Sigma} \mathbf{V}^T$ is the skinny SVD of $\mathbf{Y}$, here $\mathbf{Y}=\mathbf{A}^{(v)}+\mu_3^{-1}{\mathbf{\Lambda}_3^{(v)}}$. $\pi_\beta(\mathbf{\mathbf{\Sigma}})$ denotes soft thresholding operator defined as $\pi_\beta(\mathbf{\Sigma})=(|\mathbf{\Sigma}|-\beta)_+sgn(\mathbf{\Sigma})$ and $t_+=max(0,t)$.

\textbf{Update rule for $\mathbf{C}_2^{(v)}$ at iteration $k+1$.} Given $ \mathbf{A}^{(v)}$ at iteration $k+1$ and $\mathbf{\Lambda}_2^{(v)}$ at iteration $k$, 
we minimize the $\mathcal{L}$ in equation (\ref{Lag}) with respect to $\mathbf{C}_2^{(v)}$:
\begin{equation} \label{Lag_C2}
\begin{split}
\min_{\mathbf{C}_2^{(v)}}\mathcal{L}\big(\mathbf{C}_2^{(v)},\mathbf{A}^{(v)},{\mathbf{\Lambda}_2^{(v)}}\big)
&=\min_{\mathbf{C}_2^{(v)}}\beta_2\big\|\mathbf{C}_2^{(v)}\big\|_1+\frac{\mu_2}{2}\Big\|\mathbf{A}^{(v)}-\mathbf{C}_2^{(v)}+\frac{\mathbf{\Lambda}_2^{(v)}}{\mu_2}\Big\|_F^2\\
{\mathbf{C}_2^{(v)}} &= {\mathbf{C}_2^{(v)}}-diag({\mathbf{C}_2^{(v)}}).
\end{split}
\end{equation}
The minimization of (\ref{Lag_C2}) gives the following update rules for matrix $\mathbf{C}_2^{(v)}$ \cite{Donoho95, Daubechies04}:
\begin{equation} \label{C2_pairwise_update}
\begin{split}
\mathbf{C}_2^{(v)}&=\pi_{\frac{\beta_2}{\mu_2}}\Big(\mathbf{A}^{(v)}+\frac{{\mathbf{\Lambda}_2^{(v)}}}{\mu_2}\Big)\\
\mathbf{C}_2^{(v)}&=\mathbf{C}_2^{(v)}-diag(\mathbf{C}_2^{(v)}),
\end{split}
\end{equation}
where $\pi_\beta$ denotes soft thresholding operator applied entry-wise to $\big(\mathbf{A}^{(v)}+\mu_2^{-1}{\mathbf{\Lambda}_2^{(v)}}\big)$.

\textbf{Update rule for $\mathbf{C}_3^{(v)}$ at iteration $k+1$.} Given $ \mathbf{A}^{(v)}$ at iteration $k+1$ and $\mathbf{\Lambda}_4^{(v)}$, $\sum_{1\leq w\leq n_v, w\neq v}\mathbf{C}^{(w)}$ at iteration $k$, we minimize the objective in equation (\ref{Lag}) with respect to $\mathbf{C}_3^{(v)}$:
\begin{equation} \label{Lag_C3}
\begin{split}
\min_{\mathbf{C}_3^{(v)}}\mathcal{L}\big(\mathbf{C}_3^{(v)},\mathbf{A}^{(v)},{\mathbf{\Lambda}_4^{(v)}}\big)&=\min_{\mathbf{C}_3^{(v)}}\lambda^{(v)}\sum_{1\leq w\leq n_v, w\neq v}\big\|\mathbf{C}_3^{(v)}-\mathbf{C}^{(w)}\big\|_F^2\\&+\frac{\mu_4}{2}\big\|\mathbf{A}^{(v)}-\mathbf{C}_3^{(v)}\big\|_F^2+tr\big[{\mathbf{\Lambda}_4^{(v)}}^T\big(\mathbf{A}^{(v)}-\mathbf{C}_3^{(v)}\big)\big].
\end{split}
\end{equation}
The partial derivative of $\mathcal{L}$ in equation (\ref{Lag}) with respect to $\mathbf{C}_3^{(v)}$:
\begin{equation} \label{C3_partial}
\begin{split}
\frac{\partial\mathcal{L}}{\partial\mathbf{C}_3^{(v)}} 
&=\big[2\lambda^{(v)}(n_v-1)+\mu_4\big]\mathbf{C}_3^{(v)}-2\lambda^{(v)}\sum_{1\leq w\leq n_v, w\neq v}\mathbf{C}^{(w)}-\mu_4\mathbf{A}-\mathbf{\Lambda}_4^{(v)}.
\end{split}
\end{equation}
Setting the partial derivative in (\ref{C3_partial}) to zero:
\begin{equation} \label{C3_pairwise_update}
\mathbf{C}_3^{(v)}={\big[2\lambda^{(v)}(n_v-1)+\mu_4\big]}^{-1}\big(2\lambda^{(v)}\sum_{1\leq w\leq n_v, w\neq v}\mathbf{C}^{(w)}+\mu_4\mathbf{A}^{(v)}+{\mathbf{\Lambda}_4^{(v)}}\big).
\end{equation}

\textbf{Update rules for dual variables ${\{\mathbf{\Lambda}_i^{(v)}\}}_{i=1}^4$ at iteration $k+1 $}. Given $ \mathbf{A}^{(v)}, {\{\mathbf{C}_i^{(v)}\}}_{i=1}^3$ at iteration $k+1$, dual variables are updated with the following equations:
\begin{equation}\label{dual_var_pairwise_update}
\begin{split}
{\mathbf{\Lambda}_1^{(v)}} &= {\mathbf{\Lambda}_1^{(v)}}+\mu_1\big(\mathbf{X}^{(v)}-\mathbf{X}^{(v)}\mathbf{A}^{(v)})\\
{\mathbf{\Lambda}_2^{(v)}}&={\mathbf{\Lambda}_2^{(v)}}+\mu_2\big(\mathbf{A}^{(v)}-\mathbf{C}_2^{(v)}\big)\\
{\mathbf{\Lambda}_3^{(v)}}&={\mathbf{\Lambda}_3^{(v)}}+\mu_3\big(\mathbf{A}^{(v)}-\mathbf{C}_1^{(v)}\big)\\
{\mathbf{\Lambda}_4^{(v)}}&={\mathbf{\Lambda}_4^{(v)}}+\mu_4\big(\mathbf{A}^{(v)}-\mathbf{C}_3^{(v)}\big).
\end{split}
\end{equation}

If data is contaminated by noise and does not perfectly lie in the union of subspaces, we modify the objective function as follows:
\begin{equation} \label{corrupt_obj}
\begin{split}
\min_{\mathbf{C}^{(1)},\mathbf{C}^{(2)},...,\mathbf{C}^{(n_v)}} &\sum_{v=1}^{n_v} \Big(\frac{1}{2}\big\|\mathbf{X}^{(v)}-\mathbf{X}^{(v)}\mathbf{C}^{(v)}\big\|_F^2+\beta_1\big\|\mathbf{C}^{(v)}\big\|_*+\beta_2\big\|\mathbf{C}^{(v)}\big\|_1\Big)\\&+\sum_{1\leq v,w\leq n_v, v\neq w}\lambda^{(v)}\big\|\mathbf{C}^{(v)}-\mathbf{C}^{(w)}\big\|_F^2
\\& s.t. \quad diag(\mathbf{C}^{(v)})=\mathbf{0},\quad v=1,...n_v.
\end{split}
\end{equation}

\textbf{Update rule for $\mathbf{A}^{(v)}$ at iteration $k+1$ for corrupted data.}
Given ${\{\mathbf{C}_i^{(v)}\}}_{i=1}^3, {\{\mathbf{\Lambda}_i^{(v)}\}}_{i=1}^4$ at iteration $k$, the matrix $\mathbf{A}^{(v)}$ is obtained by equating to zero partial derivative of the augmented Lagrangian of problem (\ref{corrupt_obj}):
\begin{equation} \label{A_noisy_pairwise_update}
\begin{split}
\mathbf{A}^{(v)}=&\big[{\mathbf{X}^{(v)}}^T{\mathbf{X}^{(v)}}+(\mu_2+\mu_3+\mu_4)\mathbf{I}\big]^{-1}\times\\&\big({\mathbf{X}^{(v)}}^T{\mathbf{X}^{(v)}}+\mu_2\mathbf{C}_2^{(v)}+\mu_3\mathbf{C}_1^{(v)}+\mu_4\mathbf{C}_3^{(v)}-{\mathbf{\Lambda}_2^{(v)}}-{\mathbf{\Lambda}_3^{(v)}}-{\mathbf{\Lambda}_4^{(v)}}\big).
\end{split}
\end{equation}

Update rules for ${\{\mathbf{C}_i^{(v)}\}}_{i=1}^3$ and dual variables ${\{\mathbf{\Lambda}_i^{(v)}\}}_{i=2}^4$ are the same as in (\ref{C1_pairwise_update}), (\ref{C2_pairwise_update}), (\ref{C3_pairwise_update}), (\ref{dual_var_pairwise_update}), respectively.

These update steps are then repeated until the convergence or until the maximum number of iteration is reached. We check the convergence by verifying the following constraints at each iteration $k$: $\big\|\mathbf{A}^{(v)}-\mathbf{C}^{(v)}_1\big\|_{\infty}\leq\epsilon$, $\big\|\mathbf{A}^{(v)}-\mathbf{C}^{(v)}_2\big\|_{\infty}\leq\epsilon$, $\big\|\mathbf{A}^{(v)}-\mathbf{C}^{(v)}_3\big\|_{\infty}\leq\epsilon$ and $\big\|\mathbf{A}^{(v)}_k-\mathbf{A}^{(v)}_{k-1}\big\|_{\infty}\leq\epsilon$, for $v=1,...,n_v$. After obtaining representation matrix for each view $\allowbreak\big\{\mathbf{C}^{(1)},\mathbf{C}^{(2)},...,\mathbf{C}^{(n_v)}\big\}$, we combine them by taking the element-wise average across all views. The next step of the algorithm is to find the assignment of the data points to corresponding clusters by applying spectral clustering algorithm to the joint affinity matrix $\mathbf{W}=|\mathbf{C}_{avg}|+|\mathbf{C}_{avg}|^T$. Algorithm 1 summarizes the steps of the pairwise MLRSSC. Due to the practical reasons, we use the same initial values of $\{\mu_i\}_{i=1}^{4}$, $\rho$ and $\mu^{max}$ for different views $v$ and update $\{\mu_i\}_{i=1}^{4}$ after the optimizations of all views. However, it is possible to have more general approach with different initial values of $\{\mu_i\}_{i=1}^{4}$, $\rho$ and $\mu^{max}$ for each view $v$, but this significantly increases the number of variables for optimization.

The problem in (\ref{obj_funct_all_pair}) is convex subject to linear constraints and all its subproblems can be solved exactly. Hence, theoretical results in \cite{Hong17} guarantee the global convergence of ADMM. The computational complexity of Algorithm 1 is $O(Tn_vN^3)$, where $T$ is the number of iterations, $n_v\ll N$ is the number of views and $N$ is the number of data points. In the experiments, we set the maximal $T$ to $100$, but the algorithm converged before the maximal number of iterations is exceeded ($T\approx 15-20$).
Importantly, the computational complexity of spectral clustering step is $O(N^3)$, so the computational cost of the proposed representation learning step is $Tn_v$ times higher. 
 
\subsection{Centroid-based Multi-view Low-rank Sparse Subspace Clustering}

In addition to the pairwise MLRSSC, we also introduce objective for the centroid-based MLRSSC which enforces view-specific representations towards a common centroid. We propose to solve the following minimization problem:
\begin{equation} \label{obj_centroid}
\begin{split}
\min_{\mathbf{C}^{(1)},\mathbf{C}^{(2)},..,\mathbf{C}^{(n_v)}} \sum_{v=1}^{n_v} \big( \beta_1\big\|\mathbf{C}^{(v)}\big\|_*+\beta_2\big\|\mathbf{C}^{(v)}\big\|_1+\lambda^{(v)}\big\|\mathbf{C}^{(v)}-\mathbf{C}^*\big\|_F^2 \big)
\\ s.t.\quad \mathbf{X}^{(v)}=\mathbf{X}^{(v)}\mathbf{C}^{(v)},\quad diag(\mathbf{C}^{(v)})=\mathbf{0},\quad v=1,...n_v,
\end{split}
\end{equation}
where $\mathbf{C}^*$ denotes consensus variable.

\begin{table}[h]
\begin{singlespace}
\begin{center} \label{pairwise_MLRSSC_algorithm}
\fontsize{10}{11}\selectfont
\begin{tabular}{ l } 
\Xhline{2.5\arrayrulewidth}
 \textbf{Algorithm 1 Pairwise MLRSSC} \\ 
\hline
\textbf{Input: } $\mathbf{X}=\{\mathbf{X}^{(v)}\}_{v=1}^{n_v}$, $k$, $\beta_1$, $\beta_2$, $\{\lambda^{(v)}\}_{v=1}^{n_v},\{\mu_i,\}_{i=1}^{4}, \mu^{max}, \rho$\\ 
\textbf{Output: }Assignment of the data points to $k$ clusters\\
1: Initialize: ${\{\mathbf{C}_i^{(v)}=\mathbf{0}\}}_{i=1}^3$, $\mathbf{A}^{(v)}=\mathbf{0}$, ${\{\mathbf{\Lambda}_i^{(v)}=\mathbf{0}\}}_{i=1}^4$, $i=1,..., n_v$\\
2: \textbf{while} not converged \textbf{do}\\
3: \qquad\textbf{ for} $v=1$ \textbf{to} $n_v$ \textbf{do}\\
4: \qquad\qquad Fix others and update $\mathbf{\mathbf{A}}^{(v)}$ by solving (\ref{A_pairwise_update}) in the case of clean data\\ 
 \qquad\qquad\quad\space or (\ref{A_noisy_pairwise_update}) in the case of corrupted data\\
5: \qquad\qquad Fix others and update $\mathbf{\mathbf{C}}_1^{(v)}$ by solving (\ref{C1_pairwise_update})\\
6: \qquad\qquad Fix others and update $\mathbf{\mathbf{C}}_2^{(v)}$ by solving (\ref{C2_pairwise_update})\\
7: \qquad\qquad Fix others and update $\mathbf{\mathbf{C}}_3^{(v)}$ by solving (\ref{C3_pairwise_update})\\
8: \qquad\qquad Fix others and update dual variables $\mathbf{\Lambda}_2^{(v)},\mathbf{\Lambda}_3^{(v)},\mathbf{\Lambda}_4^{(v)}$ by solving (\ref{dual_var_pairwise_update})\\ 
\qquad\qquad\quad\space and also $\mathbf{\Lambda}_1^{(v)}$ in the case of clean data \\
9: \qquad \textbf{ end for}\\ 
10: \qquad Update $\mu_i=\min(\rho\mu_i, \mu^{max})$, $i=1,...,4$\\
11: \textbf{end while}\\
12: Combine $\big\{\mathbf{C}^{(1)},\mathbf{C}^{(2)},...,\mathbf{C}^{(n_v)}\big\}$ by taking the element-wise average\\
13: Apply spectral clustering \cite{Ng01} to the affinity matrix $\mathbf{W}=|\mathbf{C}_{avg}|+|\mathbf{C}_{avg}|^T$\\
\Xhline{2.5\arrayrulewidth}
\end{tabular}
\end{center}
\end{singlespace}
\end{table}

This objective function can be minimized by the alternating minimization cycling over the views and consensus variable. Specifically, the following two steps are repeated: (1) fix consensus variable $\mathbf{C}^*$ and update each $\mathbf{C}^{(v)}$, $v=1,...,n_v$ while keeping all others fixed and (2) fix $\mathbf{C}^{(v)}, v=1,...,n_v$ and update $\mathbf{C}^*$.

By fixing all variables except one $\mathbf{C}^{(v)}$, we solve the following problem:
\begin{equation}
\begin{split}
\min_{\mathbf{C}^{(v)}} \beta_1\big\|\mathbf{C}^{(v)}\big\|_*+\beta_2\big\|\mathbf{C}^{(v)}\big\|_1+\lambda^{(v)}\big\|\mathbf{C}^{(v)}-\mathbf{C}^*\big\|_F^2 
\\ s.t.\quad \mathbf{X}^{(v)}=\mathbf{X}^{(v)}\mathbf{C}^{(v)},\quad diag(\mathbf{C}^{(v)})=\mathbf{0}.
\end{split}
\end{equation}
Again, we solve the convex optimization problem using ADMM. We introduce auxiliary variables $\mathbf{C}_1^{(v)}$, $\mathbf{C}_2^{(v)}$, $\mathbf{C}_3^{(v)}$ and $\mathbf{A}^{(v)}$ and reformulate the original problem:
\begin{equation}
\begin{split}
&\min_{\mathbf{C}_1^{(v)},\mathbf{C}_2^{(v)},\mathbf{C}_3^{(v)},\mathbf{A}^{(v)}} \beta_1\big\|\mathbf{C}_1^{(v)}\big\|_*+\beta_2\big\|\mathbf{C}_2^{(v)}\big\|_1+\lambda^{(v)}\big\|\mathbf{C}_3^{(v)}-\mathbf{C}^*\big\|_F^2
\\ s.t.\quad \mathbf{X}^{(v)}&=\mathbf{X}^{(v)}\mathbf{A}^{(v)},\quad \mathbf{A}^{(v)}=\mathbf{C}_2^{(v)}-diag(\mathbf{C}_2^{(v)}),\quad \mathbf{A}^{(v)}=\mathbf{C}_1^{(v)}, \quad \mathbf{A}^{(v)}=\mathbf{C}_3^{(v)}.
\end{split}
\end{equation}
The augmented Lagrangian is:
\begin{equation} \label{Lag_centroid}
\begin{split}
\mathcal{L}&\big({\{{\mathbf{C}_i^{(v)}}\}}_{i=1}^3,\mathbf{A}^{(v)},{\{\mathbf{\Lambda}_i^{(v)}\}}_{i=1}^4\big) = \beta_1\big\|\mathbf{C}_1^{(v)}\big\|_*+\beta_2\big\|\mathbf{C}_2^{(v)}\big\|_1+\lambda^{(v)}\big\|\mathbf{C}_3^{(v)}-\mathbf{C}^*\big\|_F^2\\&+ \frac{\mu_1}{2}\big\|\mathbf{X}^{(v)}-\mathbf{X}^{(v)}\mathbf{A}^{(v)}\big\|_F^2
+\frac{\mu_2}{2}\big\|\mathbf{A}^{(v)}-\mathbf{C}_2^{(v)}+diag(\mathbf{C}_2)\big\|_F^2+\frac{\mu_3}{2}\big\|\mathbf{A}^{(v)}-\mathbf{C}_1\big\|_F^2
\\&+\frac{\mu_4}{2}\big\|\mathbf{A}^{(v)}-\mathbf{C}_3^{(v)}\big\|_F^2
+tr\big[{\mathbf{\Lambda}_1^{(v)}}^T(\mathbf{X}^{(v)}-\mathbf{X}^{(v)}\mathbf{A}^{(v)})\big]+tr\big[{\mathbf{\Lambda}_3^{(v)}}^T(\mathbf{A}^{(v)}-\mathbf{C}_1^{(v)})\big]\\&+tr\big[{\mathbf{\Lambda}_2^{(v)}}^T(\mathbf{A}^{(v)}-\mathbf{C}_2^{(v)}+diag(\mathbf{C}_2^{(v)})\big]+tr\big[{\mathbf{\Lambda}_4^{(v)}}^T(\mathbf{A}^{(v)}-\mathbf{C}_3^{(v)})\big].
\end{split}
\end{equation}
\textbf{Update rule for $\mathbf{C}_3^{(v)}$ at iteration $k+1$.} Given $ \mathbf{A}^{(v)}$ at iteration $k+1$ and $\mathbf{C}^*$, $\mathbf{\Lambda}_4^{(v)}$ at iteration $k$, minimization of the objective in equation (\ref{Lag_centroid}) with respect to $\mathbf{C}_3^{(v)}$ leads to the following update rule for $\mathbf{C}_3^{(v)}$:
\begin{equation} \label{C3_centroid_update}
\mathbf{C}_3^{(v)}=\big(2\lambda^{(v)}+\mu_4\big)^{-1}\big(2\lambda^{(v)}\mathbf{C}^*+\mu_4\mathbf{A}^{(v)}+{\mathbf{\Lambda}_4^{(v)}}\big).
\end{equation}

\textbf{Update rule for $\mathbf{C}^*$.} By setting the partial derivative of the objective function in equation (\ref{obj_centroid}) with respect to $\mathbf{C}^*$ to zero we get the closed-form solution to $\mathbf{C}^*$:
\begin{equation} \label{C*_update}
\mathbf{C}^*=\frac{\sum_{v=1}^{n_v}\lambda^{(v)}\mathbf{C}^{(v)}}{\sum_{v=1}^{n_v}\lambda^{(v)}}.
\end{equation}
It is easy to check that update rules for variables $\mathbf{A}^{(v)}$, $\mathbf{C}_1^{(v)}$, $\mathbf{C}_2^{(v)}$ and dual variables ${\{\mathbf{\Lambda}_i^{(v)}\}}_{i=1}^4$ are the same as in the pairwise similarities based multi-view LRSSC (equations (\ref{A_pairwise_update}), (\ref{C1_pairwise_update}),(\ref{C2_pairwise_update}) and (\ref{dual_var_pairwise_update})). 

In order to extend the model to the data contaminated by additive white Gaussian noise, the objective in (\ref{obj_centroid}) is modified as follows:
\begin{equation}
\begin{split}
\min_{\mathbf{C}^{(1)},...,\mathbf{C}^{(n_v)}} \sum_{v=1}^{n_v} \frac{1}{2}\big\|\mathbf{X}^{(v)}&-\mathbf{X}^{(v)}\mathbf{C}^{(v)}\big\|_F^2+\beta_1\big\|\mathbf{C}^{(v)}\big\|_*+\beta_2\big\|\mathbf{C}^{(v)}\big\|_1+\lambda^{(v)}\big\|\mathbf{C}^{(v)}-\mathbf{C}^*\big\|_F^2 
\\& s.t. \quad diag(\mathbf{C}^{(v)})=\mathbf{0},\quad v=1,...n_v.
\end{split}
\end{equation}
Compared to the model for clean data, the only update rule that needs to be modified is for $\mathbf{A}^{(v)}$, which is the same as in pairwise MLRSSC given in equation (\ref{A_noisy_pairwise_update}).

In centroid-based MLRSSC there is no need to combine affinity matrices across views, since the joint affinity matrix can be directly computed from the centroid matrix i.e. $\mathbf{W}=|\mathbf{C}^{*}|+|\mathbf{C}^{*}|^T$.
Algorithm 2 summarizes the steps of centroid-based MLRSSC. The computational complexity of Algorithm 2 is the same as the complexity of Algorithm 1.

\begin{table}[h]
\begin{singlespace}
\begin{center} \label{centroid_MLRSSC_algorithm}
\fontsize{10}{11}\selectfont
\begin{tabular}{ l } 
\Xhline{2.5\arrayrulewidth}
 \textbf{Algorithm 2 Centroid-based MLRSSC} \\ 
\hline
\textbf{Input: } $\mathbf{X}=\{\mathbf{X}^{(v)}\}_{v=1}^{n_v}$, $k$, $\beta_1$, $\beta_2$, $\{\lambda^{(v)}\}_{v=1}^{n_v},\{\mu_i,\}_{i=1}^{4}, \mu^{max}, \rho$\\ 
\textbf{Output: }Assignment of the data points to $k$ clusters\\
1: Initialize: ${\{\mathbf{C}_i^{(v)}=\mathbf{0}\}}_{i=1}^3$, $\mathbf{C}^*=\mathbf{0}$, $\mathbf{A}^{(v)}=\mathbf{0}$, ${\{\mathbf{\Lambda}_i^{(v)}=\mathbf{0}\}}_{i=1}^4$, $i=1,..., n_v$\\
2: \textbf{while} not converged \textbf{do}\\
3: \qquad\textbf{ for} $v=1$ \textbf{to} $n_v$ \textbf{do}\\
4: \qquad\qquad Fix others and update $\mathbf{\mathbf{A}}^{(v)}$ by solving (\ref{A_pairwise_update}) in the case of clean data\\ 
 \qquad\qquad\quad\space or (\ref{A_noisy_pairwise_update}) in the case of corrupted data\\
5: \qquad\qquad Fix others and update $\mathbf{\mathbf{C}}_1^{(v)}$ by solving (\ref{C1_pairwise_update})\\
6: \qquad\qquad Fix others and update $\mathbf{\mathbf{C}}_2^{(v)}$ by solving (\ref{C2_pairwise_update})\\
7: \qquad\qquad Fix others and update $\mathbf{\mathbf{C}}_3^{(v)}$ by solving (\ref{C3_centroid_update})\\
8: \qquad\qquad Fix others and update dual variables $\mathbf{\Lambda}_2^{(v)},\mathbf{\Lambda}_3^{(v)},\mathbf{\Lambda}_4^{(v)}$ by solving (\ref{dual_var_pairwise_update})\\ 
\qquad\qquad\quad\space and also $\mathbf{\Lambda}_1^{(v)}$ in the case of clean data \\
9: \qquad \textbf{ end for}\\
10: \qquad Update $\mu_i=\min(\rho\mu_i, \mu^{max})$, $i=1,...,4$\\
11: \qquad Fix others and update centroid $\mathbf{\mathbf{C}}^*$ by solving (\ref{C*_update})\\ 
12: \textbf{end while}\\
13: Apply spectral clustering \cite{Ng01} to the affinity matrix $\mathbf{W}=|\mathbf{C}^{*}|+|\mathbf{C}^{*}|^T$\\
\Xhline{2.5\arrayrulewidth}
\end{tabular}
\end{center}
\end{singlespace}
\end{table}

\section{Kernel Multi-view Low-rank Sparse Subspace Clustering}

The spectral decomposition of Laplacian enables spectral clustering to separate data points with nonlinear hypersurfaces. However, by representing data points as a linear combination of other data points, the MLRSSC algorithm learns the affinity matrix that models the linear subspace structure of the data. In order to recover nonlinear subspaces, we propose to solve the MLRSSC in RKHS by implicitly mapping data points into a high dimensional feature space.

We define $\Phi: \mathcal{X}\rightarrow \mathcal{F}$ to be a function that maps the original input space $\mathcal{X}$ to a high (possibly infinite) dimensional feature space $\mathcal{F}$. Since the presented update rules for the corrupted data of both pairwise and centroid-based MLRSSC depend only on the dot products $\big\langle \mathbf{X}^{(v)},{\mathbf{X}^{(v)}}\big\rangle={\mathbf{X}^{(v)}}^T\mathbf{X}^{(v)}, v=1,...,n_v$, both approaches can be solved in RKHS and extended to model nonlinear manifold structure. 

Let $\Phi(\mathbf{X}^{(v)})=\big\{\Phi(\textbf{x}_i^{(v)})\in \mathcal{F}\big\}_{i=1}^N$ denote the set of data points $\mathbf{X}^{(v)}=\big\{\textbf{x}_i^{(v)}\in  {\rm I\!R}^D\big\}_{i=1}^N$ mapped into high-dimensional feature space. The objective function of pairwise kernel MLRSSC for data contaminated by noise is the following:
\begin{equation}
\begin{split}
\min_{\mathbf{C}^{(1)},\mathbf{C}^{(2)},...,\mathbf{C}^{(n_v)}} &\sum_{v=1}^{n_v} \Big(\frac{1}{2}\big\|\Phi(\mathbf{X}^{(v)})-\Phi(\mathbf{X}^{(v)})\mathbf{C}^{(v)}\big\|_F^2+\beta_1\big\|\mathbf{C}^{(v)}\big\|_*+\beta_2\big\|\mathbf{C}^{(v)}\big\|_1\Big)\\&+\sum_{1\leq v,w\leq n_v, v\neq w}\lambda^{(v)}\big\|\mathbf{C}^{(v)}-\mathbf{C}^{(w)}\big\|_F^2
\\& s.t. \quad diag(\mathbf{C}^{(v)})=\mathbf{0},\quad v=1,...n_v.
\end{split}
\end{equation}

Similarly, the objective function of centroid-based MLRSSC in feature space for corrupted data is:
\begin{equation}
\begin{split}
\min_{\mathbf{C}^{(1)},\mathbf{C}^{(2)},...,\mathbf{C}^{(n_v)}} &\sum_{v=1}^{n_v} \Big(\frac{1}{2}\big\|\Phi(\mathbf{X}^{(v)})-\Phi(\mathbf{X}^{(v)})\mathbf{C}^{(v)}\big\|_F^2+\beta_1\big\|\mathbf{C}^{(v)}\big\|_*+\beta_2\big\|\mathbf{C}^{(v)}\big\|_1\\&+\lambda^{(v)}\big\|\mathbf{C}^{(v)}-\mathbf{C}^*\big\|_F^2 \Big)
\\& s.t. \quad diag(\mathbf{C}^{(v)})=\mathbf{0},\quad v=1,...n_v.
\end{split}
\end{equation}

Since $\mathbf{A}^{(v)}$ is the only variable that depends on $\mathbf{X}^{(v)}$, the update rules for ${\{\mathbf{C}_i^{(v)}\}}_{i=1}^3$ and dual variables ${\{\mathbf{\Lambda}_i^{(v)}\}}_{i=2}^4$ remain unchanged.

\textbf{Update rule for $\mathbf{A}^{(v)}$ at iteration $k+1$}. Given ${\{\mathbf{C}_i^{(v)}\}}_{i=1}^3, {\{\mathbf{\Lambda}_i^{(v)}\}}_{i=2}^4$ at iteration $k$, the $\mathbf{A}^{(v)}$ is updated by the following update rule:
\begin{equation}
\begin{split}
\mathbf{A}^{(v)}=&\big[\Phi(\mathbf{X}^{(v)})^T\Phi({\mathbf{X}^{(v)}})+(\mu_2+\mu_3+\mu_4)\mathbf{I}\big]^{-1}\times\\&\big[\Phi({\mathbf{X}^{(v)}})^T\Phi({\mathbf{X}^{(v)}})+\mu_2\mathbf{C}_2^{(v)}+\mu_3\mathbf{C}_1^{(v)}+\mu_4\mathbf{C}_3^{(v)}-\mathbf{\Lambda}_2^{(v)}-\mathbf{\Lambda}_3^{(v)}-\mathbf{\Lambda}_4^{(v)}\big].
\end{split}
\end{equation}
Substituting the dot product $\big\langle\Phi(\mathbf{X}^{(v)}),\Phi(\mathbf{X}^{(v)})\big\rangle$ with the Gram matrix $\mathbf{K}^{(v)}$, we get the following update rule for $\mathbf{A}^{(v)}$:
\begin{equation}
\begin{split}
\mathbf{A}^{(v)}=&\big[\mathbf{K}^{(v)}+(\mu_2+\mu_3+\mu_4)\mathbf{I}\big]^{-1}\times\\&\big[\mathbf{K}^{(v)}+\mu_2\mathbf{C}_2^{(v)}+\mu_3\mathbf{C}_1^{(v)}+\mu_4\mathbf{C}_3^{(v)}-\mathbf{\Lambda}_2^{(v)}-\mathbf{\Lambda}_3^{(v)}-\mathbf{\Lambda}_4^{(v)}\big].
\end{split}
\end{equation}
Update rule for $\mathbf{A}^{(v)}$ is the same in pairwise and centroid-based versions of the algorithm.

\section{Experiments} 

In this section we present results that demonstrate the effectiveness of the proposed algorithms. The performance is measured on one synthetic and three real-world datasets that are commonly used to evaluate the performance of multi-view algorithms. Moreover, we introduce novel real-world multi-view dataset from molecular biology domain. We compared MLRSSC with the state-of-the-art multi-view subspace clustering algorithms, as well as with two baselines: best single view LRSSC and feature concatenation LRSSC. 
 
\subsection{Datasets}

We report the experimental results on synthetic and four real-world datasets. %\footnote{We did not perform test on commonly used Caltech-101 dataset since only kernel matrices are available.}
We give a brief description of each dataset. Statistics of the datasets are summarized in Table \ref{datasets_summary}.

\textbf{UCI Digit dataset} is available from the UCI repository\footnote{http://archive.ics.uci.edu/ml/datasets/Multiple+Features}. This dataset consists of 2000 examples of handwritten digits (0-9) extracted from Dutch utility maps. There are 200 examples in each class, each represented with six feature sets. Following experiments in \cite{Lu16}, we used three feature sets: 76 Fourier coefficients of the character shapes, 216 profile correlations and 64 Karhunen-Love coefficients.

\textbf{Reuters dataset} \cite{Lewis2004} contains features of documents available in five different languages and their translations over a common set of six categories. All documents are in the bag-of-words representation. We use documents originally written in English as one view and their translations to French, German, Spanish and Italian as four other views. We randomly sampled 100 documents from each class, resulting in a dataset of 600 documents. %, as done in \cite{Lu16}.

\textbf{3-sources dataset}\footnote{http://mlg.ucd.ie/datasets/3sources.html} is news articles dataset collected from three online news sources: BBC, Reuters, and The Guardian. All articles are in the bag-of-words representation. Of 948 articles, we used 169 that are available in all three sources. Each article in the dataset is annotated with a dominant topic class. 
  
\textbf{Prokaryotic phyla dataset} contains 551 prokaryotic species described with heterogeneous multi-view data including textual data and different genomic representations \cite{Brbic16}. Textual data consists of bag-of-words representation of documents describing prokaryotic species and is considered as one view. In our experiments we use two genomic representations: (i) the proteome composition, encoded as relative frequencies of amino acids (ii) the gene repertoire, encoded as presence/absence indicators of gene families in a genome. In order to reduce the dimensionality of the dataset, we apply principal component analysis (PCA) on each of the three views separately and retain principal components explaining $90\%$ of the variance. Each species in the dataset is labeled with the phylum it belongs to. Unlike previous datasets, this dataset is unbalanced. The most frequently occurring cluster contains $313$ species, while the smallest cluster contains $35$ species. 

\textbf{Synthetic dataset} was generated in a way described in \cite{Kumar11, Yi05}. $1000$ points are generated from two views, where data points for each view are generated from two-component Gaussian mixture models. Cluster means and covariance matrices for view $1$ are: $\mu_1^{(1)}=(1\quad1)$, $\Sigma_1^{(1)}=(1\quad0.5;\quad 0.5\quad1.5)$, $\mu_2^{(1)}=(2\quad2)$, $\Sigma_2^{(1)}=(0.3\quad0;\quad 0\quad0.6)$, and for view 2 are: $\mu_1^{(2)}=(2\quad2)$, $\Sigma_1^{(2)}=(0.3\quad0;\quad 0\quad0.6)$, $\mu_2^{(2)}=(1\quad1)$, $\Sigma_2^{(2)}=(1\quad0.5;\quad 0.5\quad1.5)$.

\begin{table}[h]
\caption{Statistics of the multi-view datasets}
\fontsize{9}{11}\selectfont
\begin{center} \label{datasets_summary} 
\begin{tabular}{ | c|c|c|c |} 
\hline 
 \textbf{Dataset} & \textbf{Samples} & \textbf{Views} & \textbf{Clusters} \\ 
 \hline
 UCI Digit & 2000 & 3 & 10 \\ 
 \hline
 Reuters & 600 & 5 & 6 \\ 
 \hline
 3-sources & 169 & 3 & 6 \\ 
 \hline
 Prokaryotic & 551 & 3 & 4 \\ 
 \hline
 Synthetic & 1000 & 2 & 2 \\ 
 \hline
\end{tabular}
%\\[8pt]
\end{center}
\end{table}

\subsection{Compared Methods and Parameters}
We compare pairwise MLRSSC, centroid-based MLRSSC and kernel extensions of both algorithms (KMLRSSC) with the best performing state-of-the-art multi-view subspace clustering algorithms, including Co-regularized Multi-view Spectral Clustering (Co-Reg) \cite {Kumar11}, Robust Multi-view Spectral Clustering (RMSC) \cite{Xia14} and Convex Sparse Multi-view Spectral Clustering (CSMSC) \cite{Lu16}. Moreover, we also compare MLRSSC algorithms with two LRSSC baselines: (i) best single view Low-rank Sparse Subspace Clustering (LRRSC) \cite{Wang13} that performs single view LRSSC on each view and takes the individual view that achieves the best performance, and (ii) feature concatenation LRRSC that concatenates features of each individual view and performs single-view LRSSC on the joint view representation.
%\begin{itemize}
%\item \textbf{Best Single View Low-rank Sparse Subspace Clustering:} performs single view LRSSC \cite{Wang13} on each view and takes the individual view that achieves the best performance.
%\item \textbf{Feature Concatenation Low-rank Sparse Subspace Clustering:} concatenates features of each individual view and performs single-view LRSSC on the joint view representation.
%\item \textbf{Co-regularized Multi-view Spectral Clustering (Co-Reg) \cite {Kumar11}:} Co-regularization approach that performs regularization on the eigenvectors of Laplacians to achieve consistency of Laplacians across the views.
%\item \textbf{Robust Multi-view Spectral Clustering (RMSC) \cite{Xia14}:} Markov chain method for multi-view spectral clustering that constructs transition probability matrix from each individual view and recovers shared transition probability matrix via low-rank and sparse decomposition.
%\item \textbf{Convex Sparse Multi-view Spectral Clustering (CSMSC) \cite{Lu16}:} Extension of convex sparse spectral clustering, proposed in \cite{Lu16}, to multi-view data. 
%\end{itemize}

Co-regularized multi-view SC has a parameter $\alpha$ that we vary from $0.01$ to $0.05$ with step $0.01$ \cite{Kumar11}. We choose $\lambda$ in RMSC from the set of the values: $\allowbreak\{0.005, 0.01, 0.05, 0.1, 0.5, 1, 5, 10, 50, 100\}$, as tested in \cite{Xia14}. Parameter $\alpha$ in CSMSC is chosen from $\{10^{-1},10^{-2}\}$ and parameter $\beta$ from $\{10^{-3},10^{-4},10^{-5}\}$ \cite{Lu16}. For all these algorithms the standard deviation of Gaussian kernel used to build similarity matrix is set to the median of the pairwise Euclidean distances between the data points \cite{Kumar11, Xia14, Lu16}. The number of iterations of the Co-Reg SC is set to $100$, but it converged within less than $10$ iterations. The number of iterations of the CSMSC is set to $200$ \cite{Lu16} and of the RMSC to $300$, as set in the available source code provided by the authors. All other parameters of these algorithms are set to values based on the respective source codes provided by their authors. 

For LRSSC and MLRSSC we first choose penalty parameter $\mu$ from the set of values $\{10^1,10^2,10^3,\\10^4\}$ with fixed $\beta_1$, $\beta_2$ and $\lambda^{(v)}$. We set the same value $\mu$ for all constraints $(\mu_i,i=1,...,4)$, but one can also optimize $\mu$ for each of the constraints. In each iteration we update $\mu$ to be $\rho \mu$ with fixed $\rho$ of $1.5$ and till the maximal value of $\mu$ (set to $10^{6}$) is reached. For single-view LRSSC $\rho$ is set to $1$. Low-rank parameter $\beta_1$ is tuned from $0.1$ to $0.9$ with step $0.2$ and sparsity parameter $\beta_2$ is set to $(1-\beta_1)$. Consensus parameter $\lambda$ is tuned from $0.3$ to $0.9$ with step $0.2$. It is also possible to use different $\lambda^{(v)}$ for each view $v$, but since we did not have any prior information about the importance of views, we use the same $\lambda=\lambda^{(v)}$ for each view $v$. For all datasets we use the variant adjusted for the corrupted data, except for the UCI digit dataset. In the kernel extension of MLRSSC, we use Gaussian kernel and optimize standard deviation for each view separately in range $\{0.5, 1, 5, 10, 50\}$ times the median of the pairwise Euclidean distances between the data points, while holding other parameters fixed. Best sigma for pairwise MLRSSC was also used for centroid MLRSSC without further optimization. The maximum number of iterations is set to 100 and the convergence error tolerance to $\epsilon=10^{-3}$ for linear MLRSSC and $\epsilon=10^{-5}$ for kernel MLRSSC. We tune the parameters of each algorithm and report the best performance.

All compared methods have k-means as the last step of the algorithm. Since k-means depends on the initial cluster centroid positions and can yield different solution with different initializations, we run k-means 20 times and report the means and standard deviations of the performance measures. We evaluate clustering performance using five different metrics: precision, recall, F-score, normalized mutual information (NMI) and adjusted rand index (Adj-RI) \cite{Manning08}. For all these metrics, the higher value indicates better performance.

\subsection{Results}

Table \ref{performance_results} compares the clustering performance of the MLRSSC with other algorithms on four real-world datasets and one synthetic dataset. 
Results indicate that MLRSSC consistently outperforms all other methods in terms of all tested measures. On all five datasets, MLRSSC improves performance to a large extent which demonstrates the importance of combined low-rank and sparsity constraints. More specifically, the average NMI of the MLRSSC is higher than the second best method by $7\%$, $9\%$, $4\%$, $12\%$ and $2\%$ on the 3-sources, Reuters, UCI digit, Prokaryotic and synthetic datasets, respectively. Similar improvements can also be observed when using other metrics for measuring clustering performance.

\begin{table} [!t]  
\caption{Performance of different algorithms on five multi-view datasets. The mean and standard deviation of 20 runs of k-means clustering algorithm with different random initializations are reported.}
\begin{center} \label{performance_results}
\resizebox{\columnwidth}{!} { 
%\afterpage
\begin{tabular}{ | c | c | c | c | c | c | c | } 
\hline
\textbf{Dataset} & \textbf{Method} & \textbf{F-score} & \textbf{Precision} & \textbf{Recall} & \textbf{NMI} & \textbf{Adj-RI} \\ \hline
	 & Best Single View LRSSC & 0.569 (0.039)	& 0.604 (0.015)	& 0.541 (0.058)	& 0.496 (0.024)	& 0.449 (0.042)\\
	 & Feat Concat LRSSC & 0.579 (0.048)	& 0.593 (0.031)	& 0.571 (0.078)	& 0.521 (0.015)	& 0.455 (0.054) \\
	 & Co-Reg Pairwise  & 0.463 (0.020) & 0.504 (0.049) & 0.437 (0.033) & 0.519 (0.036)	& 0.315 (0.033) \\ 
	& Co-reg Centroid & 0.505 (0.032) &	0.551 (0.052) &	0.467 (0.025) &	0.514 (0.026) &	0.370 (0.045) \\
	\textbf{3-sources} & RMSC & 0.477 (0.033) &	0.515 (0.034) &	0.453 (0.036) &	0.517 (0.024) &	0.330 (0.045) \\
	& CSMSC & 0.482 (0.026)	& 0.518 (0.056)	& 0.464 (0.027) & 0.518 (0.026) & 0.335 (0.039) \\
	& Pairwise MLRSSC & \textbf{0.659 (0.049)} &	\textbf{0.707 (0.051)} &	\textbf{0.619 (0.056)} & 0.594 (0.025) &	\textbf{0.565 (0.060)}\\	
	& Centroid MLRSSC & 0.654 (0.042) &	0.696 (0.055) &\textbf{	0.619 (0.052)} & \textbf{0.595 (0.021)} &	0.557 (0.053)\\	
	 & Pairwise KMLRSSC & 	0.541 (0.025) &	0.619 (0.032) &	0.482 (0.033)	& 0.529 (0.020)	& 0.424 (0.029)\\	
	 & Centroid KMLRSSC &  	0.556 (0.045) &	0.622 (0.049) &	0.503 (0.044)	& 0.533 (0.031) &	0.439 (0.056)\\ \hline
	 & Best Single View LRSSC & 0.333 (0.003)	& 0.313 (0.007)	& 0.357 (0.019)	& 0.245 (0.008)	& 0.191 (0.005) \\
	 & Feat Concat LRSSC & 0.347 (0.005)	& 0.319 (0.010)	& 0.384 (0.022)	& 0.283 (0.006)	& 0.204 (0.008) \\
	 & Co-Reg Pairwise & 0.371 (0.009)  &	0.344 (0.016) &	0.410 (0.023) &	0.300 (0.014) & 0.233 (0.017) \\
	 & Co-reg Centroid & 0.362 (0.017) & 0.331 (0.022)	& 0.409 (0.020)	& 0.291 (0.014)	& 0.221 (0.023) \\
	\textbf{Reuters} & RMSC & 0.361 (0.019) & 0.325 (0.012)	& 0.412 (0.023)	& 0.297 (0.018)	& 0.217 (0.015) \\
	 & CSMSC & 0.365 (0.005) &	0.327 (0.010) &	0.420 (0.014) &	0.295 (0.020) &	0.220 (0.008) \\
	 & Pairwise MLRSSC & 0.428 (0.012) &	0.389 (0.024) &	\textbf{0.486 (0.019)} &	0.390 (0.018) &	0.300 (0.021) \\
	 & Centroid MLRSSC & \textbf{0.432 (0.010)} & 0.395 (0.023) &	0.482 (0.025) &	\textbf{0.394 (0.015)} & 0.306 (0.017) \\
	 & Pairwise KMLRSSC & 0.429 (0.013) &	\textbf{0.415 (0.018)} &	0.446 (0.016) &	0.380 (0.018) &	\textbf{0.311 (0.017)}  \\
	 & Centroid KMLRSSC & 0.426 (0.013) & 0.410 (0.018) &	0.443 (0.015) &	0.373 (0.016) &	0.307 (0.017) \\ \hline
	 & Best Single View LRSSC & 0.702 (0.033)	& 0.659 (0.033)	& 0.755 (0.027)	& 0.754 (0.020)	& 0.666 (0.038) \\
	 & Feat Concat LRSSC & 0.698 (0.038)	& 0.671 (0.046)	& 0.728 (0.032)	& 0.751 (0.021)	& 0.663 (0.043) \\
	 & Co-Reg Pairwise & 0.694 (0.057) &	0.671 (0.068) &	0.718 (0.047) &	0.739 (0.036)	& 0.658 (0.065) \\
	& Co-reg Centroid & 0.754 (0.067) &	0.735 (0.082)  &	0.775 (0.050) &	0.783 (0.033)	& 0.726 (0.075) \\
	\textbf{UCI digit} & RMSC & 0.742 (0.070) & 0.728 (0.080) & 0.757 (0.061) & 0.778 (0.040) & 0.713 (0.079) \\
	 & CSMSC & 0.775 (0.045) &	0.725 (0.069) &	0.836 (0.015) &	0.819 (0.019) &	0.748 (0.051) \\
	& Pairwise MLRSSC & 0.830 (0.048) &	0.809 (0.070) &	0.854 (0.027) &	0.851 (0.023) &	0.810 (0.054)\\
	& Centroid MLRSSC & 0.835 (0.047) &	0.819 (0.066) &	0.854 (0.027) &	0.854 (0.023) &	0.817 (0.053)\\
	& Pairwise KMLRSSC & 0.827 (0.063) & 0.800 (0.078) & 0.861 (0.022)	& 0.855 (0.027) &	0.807 (0.072)\\
	& Centroid KMLRSSC & \textbf{0.840 (0.043)} &	\textbf{0.820 (0.065)} &	\textbf{0.862 (0.019)} &	\textbf{0.858 (0.020)} &	\textbf{0.822 (0.048)}\\ \hline
	 & Best Single View LRSSC & 0.579 (0.057) & 0.551 (0.016)	& 0.634 (0.100)	& 0.233 (0.026)	& 0.280 (0.051) \\
	 & Feat Concat LRSSC & 0.584 (0.054) & 0.542 (0.015)	& 0.644 (0.092)	& 0.218 (0.029)	& 0.275 (0.057) \\
	 & Co-Reg Pairwise & 0.468 (0.023) & 0.568 (0.023) & 0.398 (0.022) & 0.286 (0.021)	& 0.213 (0.031) \\
	 & Co-reg Centroid & 0.459 (0.010) & 0.567 (0.010) & 0.386 (0.012) & 0.296 (0.018)	& 0.206 (0.012) \\
	 \textbf{Prokaryotic} & RMSC & 0.447 (0.027)	& 0.567 (0.038)	& 0.369 (0.023)	& 0.315 (0.041)	& 0.198 (0.044) \\
	 & CSMSC & 0.462 (0.026) &	0.565 (0.024) &	0.391 (0.026) &	0.269 (0.022) &	0.206 (0.033) \\
	 & Pairwise MLRSSC & \textbf{0.591 (0.016)} & 0.624 (0.003)	& 0.566 (0.036)	& 0.322 (0.002)	& 0.345 (0.016) \\
	 & Centroid MLRSSC & 0.574 (0.028) &	0.530 (0.014) &\textbf{	0.756 (0.124)} &	0.202 (0.018)	& 0.258 (0.032)\\
	 & Pairwise KMLRSSC & \textbf{0.591 (0.056)} & \textbf{0.725 (0.068)} & 0.499 (0.048) & \textbf{0.437 (0.039)} & \textbf{0.398 (0.082)}  \\
	 & Centroid KMLRSSC & 0.582 (0.070) & 0.712 (0.079) & 0.492 (0.062) & 0.424 (0.046) & 0.384 (0.100)  \\ \hline
	  & Best Single View LRSSC  & 0.624 (0.000) &	0.560 (0.000) &	0.704 (0.000) &	0.182 (0.000)	& 0.152 (0.000) \\
	 & Feat Concat LRSSC   & 0.682 (0.000) &	0.682 (0.000) &	0.682 (0.000) &	0.283 (0.000) &	0.364 (0.000)\\
	 & Co-Reg Pairwise & 0.660 (0.000) &	0.637 (0.000) &	0.685 (0.000) &	0.260 (0.000) &	0.295 (0.000)\\
	 & Co-reg Centroid & 0.646 (0.000) &	0.630 (0.000) &	0.664 (0.000) &	0.229 (0.000) &	0.274 (0.000)\\
	 \textbf{Synthetic} & RMSC   & 0.715 (0.000) &	0.715 (0.000) &	0.715 (0.000) &	0.338 (0.000) &	0.430 (0.000)\\
	 & CSMSC & 0.730 (0.000) &	0.729 (0.000) &	0.731 (0.000) &	0.366 (0.000)	& 0.459 (0.000)\\
	 & Pairwise MLRSSC & 0.689 (0.000) &	0.689 (0.000) &	0.689 (0.000) &	0.294 (0.000) &	0.379 (0.000)\\
	 & Centroid MLRSSC  & 0.690 (0.002) &	0.690 (0.002) &	0.690 (0.002) &	0.296 (0.003) &	0.380 (0.004) \\
	 & Pairwise KMLRSSC  & 0.742 (0.000) &	0.742 (0.000) &	0.742 (0.000)	& 0.385 (0.000)	& 0.484 (0.000)\\
	 & Centroid KMLRSSC & \textbf{0.743 (0.000)} &	\textbf{0.743 (0.000)}	& \textbf{0.805 (0.000)}	& \textbf{0.388 (0.002)}	&\textbf{ 0.487 (0.000)} \\ \hline
\end{tabular}
}
\end{center}
\end{table}

Pairwise and centroid-based MLRSSC perform comparably, except on Prokaryotic dataset where pairwise MLRSSC is significantly better than the centroid-based MLRSSC, except in recall. When comparing linear MLRSSC with the kernel MLRSSC, linear MLRSSC performs better on 3-sources and Reuters datasets. %On 3-sources dataset linear MLRSSC is significantly better than the kernel MLRSSC in terms of all metrics and on Reuters dataset linear MLRSSC is significantly better or comparable to kernel MLRSSC, except in precision. 
Kernel MLRSSC outperforms linear MLRSSC on UCI Digit, Prokaryotic and synthetic datasets, although the difference on the UCI Digit dataset is not significant. %On the Prokaryotic dataset kernel MLRPPSC performs much better in terms of precision, NMI and adjusted Rand index measures. 
However, this comes with the cost of tuning more parameters for computing the kernel. Better performance of linear MLRPPSC on 3-sources and Reuters datasets is not suprising, since these datasets are very sparse (more than $95\%$ values are zeros) and have a large number of features, much higher than the number of data points. On the other hand, UCI Digit, Prokaryotic and especially synthetic datasets have dense lower-dimensional feature vectors and benefit from the projection to a high-dimensional feature space.

\subsection{Parameter Sensitivity}

MLRSSC trades-off low-rank, sparsity and consensus parameters: $\beta_1$, $\beta_2$ and $\lambda^{(v)}$, respectively. In this section, we test the effect of these parameters on the performance of the MLRSSC. In all experiments, we set the sparsity parametar $\beta_2$ to $1-\beta_1$, i.e. the higher value of a low-rank parameter leads to the lower value of a sparsity parameter and vice versa. This depends on whether the problem being solved requires exploiting more global or the local structure of the data. 

\begin{figure}[h]
		\centering
		\includegraphics[width=1.0\textwidth]{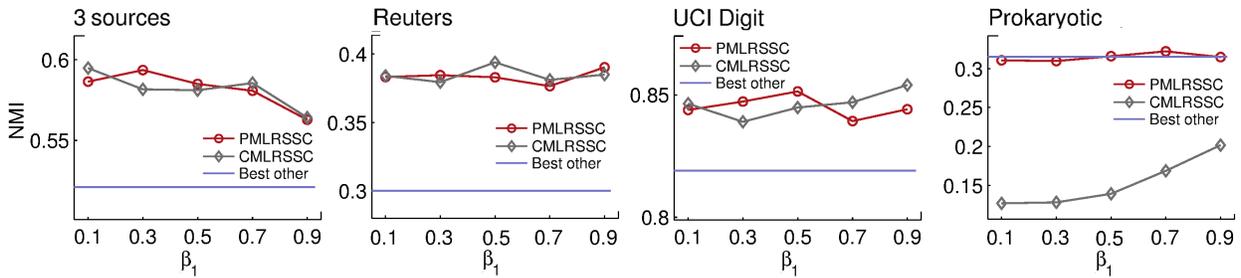}
		\caption{The performance of the MLRSSC w.r.t. NMI measure when varying low-rank parameter $\beta_1$ and keeping consensus parameter $\lambda^{(v)}$ fixed. Sparsity parameter $\beta_2$ is set to $1-\beta_1$. Blue line shows the best performing algorithm besides MLRSSC, among the algorithms listed in Table \ref{performance_results}. PMLRSSC stands for pairwise MLRSSC and CMLRSSC for centroid-based MLRSSC.} 
		\label{Fig1}
\end{figure}

Figure \ref{Fig1} shows how the NMI metrics changes with different values of low-rank parameter $\beta_1$ for both pairwise and centroid-based MLRSSC, while keeping $\lambda^{(v)}$ parameter fixed. On the 3-sources, Reuters and UCI Digit, MLRSSC algorithm outperforms the second best algorithm regardless of the choice of $\beta_1$. On the Prokaryotic dataset, pairwise MLRSSC performs comparably to RMSC, but again the algorithm is insensitive to the $\beta_1$ parameter. On the other hand, centroid-based MLRSSC lags behind on this dataset with respect to NMI measure, but consistently improves its performance with the higher values of $\beta_1$.

\begin{figure}[h]
		\centering
		\includegraphics[width=1\textwidth]{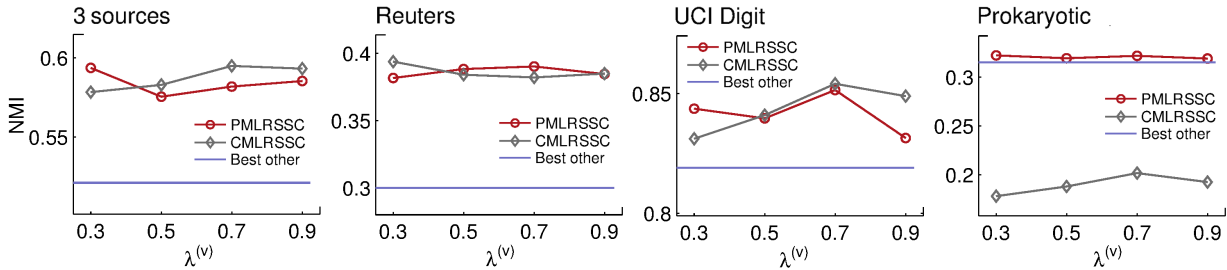}
		\caption{The performance of the MLRSSC w.r.t. NMI measure when varying consensus parameter $\lambda^{(v)}$ and keeping low-rank parameter $\beta_1$ and sparsity parameter $\beta_2$ fixed. Blue line shows the best performing algorithm besides MLRSSC, among the algorithms listed in Table \ref{performance_results}.}
		\label{Fig2}
\end{figure}

Next, we vary consensus parameter $\lambda^{(v)}$ and keep the low-rank parameter $\beta_1$ and sparsity parameter $\beta_2$ fixed. Figure \ref{Fig2} shows the performance of the MLRSSC with respect to NMI measure for different values of $\lambda^{(v)}$. Similarly as when varying $\beta_1$ parameter, the MLRSSC performs consistently better than other algorithms regardless of the choice of $\lambda^{(v)}$. Again, the only exception is the centroid-based MLRSSC on the Prokaryotic dataset. These results prove that MLRSSC is pretty stable regardless of the choice of its parameters $\beta_1$, $\beta_2$ and $\lambda^{(v)}$, as long as the parameters are chosen in an appropriate range.

 \subsection{Computational Time and Convergence}
\begin{figure}[h]
 		\centering
 		\includegraphics[width=0.65\textwidth]{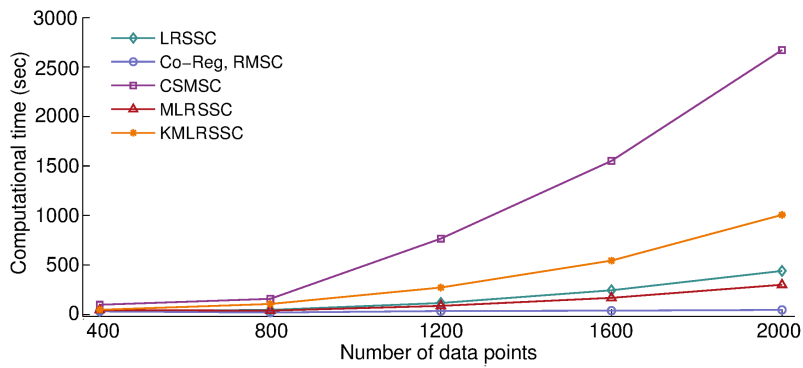}
 		\caption{Average computational time in seconds as a function of the number of data points, measured on the UCI Digit dataset. For the Co-Reg and MLRSSC algorithm times for pairwise regularization are shown, but they are similar for the centroid regularization. The difference between Co-Reg and RMSC can not be seen on this scale, so these two algorithms are shown together.}
 		\label{Fig3}
 \end{figure}

 In order to check how computational time of the MLRSSC scales with the increase of the number of data points, we perform experiments on the UCI digit dataset and compare MLRSSC with other algorithms. Computational time depends on the number of iterations and convergence conditions. We use the same number of iterations and error tolerance as when comparing performance of the algorithms. 
 Figure \ref{Fig3} shows the computational time averaged over 10 runs as a function of the number of data points. Figure demonstrates that MLRSSC is more efficient than CSMSC. Compared to Co-Reg SC and RMSC, the better performance of MLRSSC comes with a higher computational cost. 
 
\begin{figure}[H]
 		\centering
 		\includegraphics[width=1.0\textwidth]{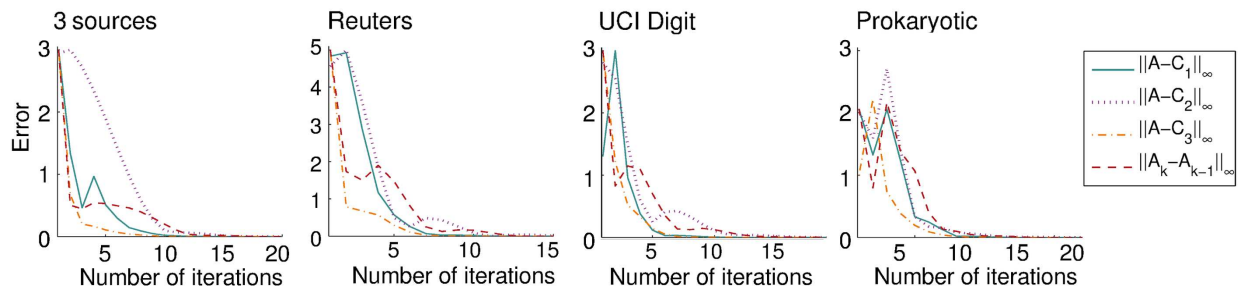}
 		\caption{Sum of normalized errors across views for pairwise MLRSSC. Behavior is very similar for centroid MLRSSC.}
 		\label{Fig4}
 \end{figure}

\begin{figure}[H]
  		\centering
  		\includegraphics[width=1.0\textwidth]{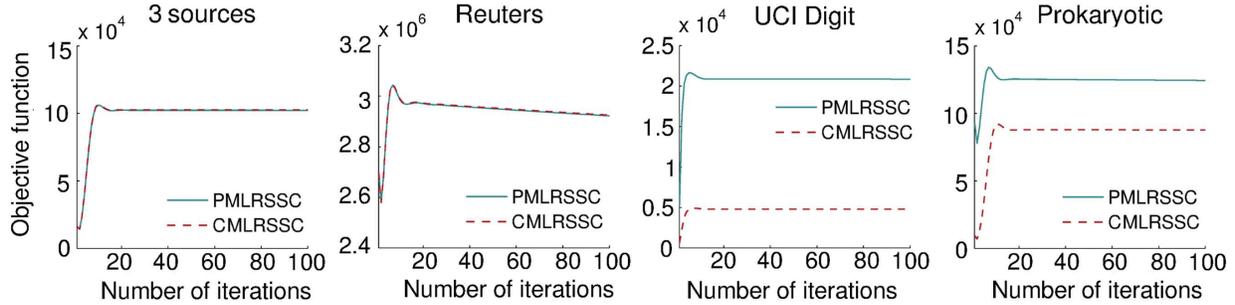}
  		\caption{Objective function value with the respect to number of iterations for pairwise and centroid MLRSSC.}
 		\label{Fig5}
 \end{figure}

 Figure \ref{Fig4} demonstrates the behavior of convergence conditions for pairwise MLRSSC. For ease of illustration, the errors are normalized and summed across views. It can be seen that on all four real-world datasets, the algorithm converges within 20 iterations.  Centroid MLRSSC exhibits very similar behavior. Figure \ref{Fig5} shows objective function value for both pairwise and centroid MLRRSC with the respect to number of iterations.
 
\section{Concluding Remarks}
In this paper we proposed multi-view subspace clustering algorithm, called Multi-view Low-rank Sparse Subspace Clustering (MLRSSC), that learns a joint subspace representation across all views. The main property of the algorithm is to jointly learn an affinity matrix constrained by sparsity and low-rank. We defined optimization problems and derived ADMM-based algorithms for pairwise and centroid-based regularization schemes. In addition, we extended the proposed MLRSSC algorithm to nonlinear subspaces by solving the related optimization problem in reproducing kernel Hilbert space. Experimental results on multi-view datasets from various domains showed that proposed algorithms outperforms state-of-the-art multi-view subspace clustering algorithms.

High computational complexity presents serious drawback of spectral clustering algorithms. In the future work, we plan to explore how to improve the efficiency of the proposed approach to be applicable to large-scale multi-view problems. Moreover, we may consider how to extend the MLRSSC algorithm to handle incomplete data.

 \section*{Acknowledgements}
This work has been supported by the Croatian Science Foundation grant IP-2016-06-5235 (Structured Decompositions of Empirical Data for Computationally-Assisted
Diagnoses of Disease) and by the Croatian Science Foundation grant HRZZ-9623 (Descriptive Induction).

%\section*{References}
 %\label{}

 \bibliographystyle{elsarticle-num} 
 \bibliography{library}
% \biboptions{sort&compress}

\end{document}